\documentclass[runningheads]{llncs}
\usepackage[T1]{fontenc}
\usepackage{graphicx}
\usepackage{booktabs}
\usepackage{multirow}
\usepackage[normalem]{ulem}
\useunder{\uline}{\ul}{}
\usepackage{hyperref}
\usepackage[accsupp]{axessibility}
\usepackage{orcidlink}
\usepackage{subcaption}

\begin{document}
\title{Greit-HRNet: Grouped Lightweight High-Resolution Network for Human Pose Estimation}

\titlerunning{Greit-HRNet}
%
%
\author{Junjia Han\inst{1}\orcidlink{0009-0002-5102-1824} \and
Yanxia Wang\inst{2}}
\authorrunning{H. Author \& W. Coauthor}
%
\institute{The University of Hong Kong, Pok Fu Lam, Hong Kong SAR, China \email{hanlonegen@gmail.com}\and Chongqing Normal University, Chongqing 400047, China \email{wangyanxia@cqnu.edu.cn}}
\maketitle              

\begin{abstract}
  As multi-scale features are necessary for human pose estimation tasks, high-resolution networks are widely applied. 
  To improve efficiency, lightweight modules are proposed to replace costly point-wise convolutions in high-resolution networks, including channel weighting and spatial weighting methods. 
  However, they fail to maintain the consistency of weights and capture global spatial information. 
  To address these problems, we present a Grouped lightweight High-Resolution Network (Greit-HRNet), in which we propose a Greit block including a group method Grouped Channel Weighting (GCW) and a spatial weighting method Global Spatial Weighting (GSW). 
  GCW modules group conditional channel weighting to make weights stable and maintain the high-resolution features with the deepening of the network, while GSW modules effectively extract global spatial information and exchange information across channels. 
  In addition, we apply the Large Kernel Attention (LKA) method to improve the whole efficiency of our Greit-HRNet. 
  Our experiments on both MS-COCO and MPII human pose estimation datasets demonstrate the superior performance of our Greit-HRNet, outperforming other state-of-the-art lightweight networks.
  \keywords{Human pose estimation \and High-resolution networks \and Efficient networks}
\end{abstract}
\section{Introduction}
\label{sec:intro}

As a basic task of pattern recognition, human pose estimation is a trending topic in computer vision. 
The key to it is to detect the key points of the human body that can effectively represent the human pose. \cite{zheng2023deep}
To achieve high performance and accuracy in detecting the key points, high-resolution feature extraction has been emphasized in recent works \cite{sigal2021human,luo2022fastnet,rui2023edite,neff2021efficienthrnet,wang2022lite,li2022dite}.

To maintain the high-resolution representation during the process of human pose estimation, HRNet \cite{sun2019deep} is introduced as an effective method, in which four parallel branches with multi-scale features are adopted, but this design causes the problem that the model is difficult to achieve high efficiency. 
To reduce the computational complexity, Small HRNet \cite{wang2020deep} is proposed with less width and depth as an efficient model while it compromises the performance and accuracy. 
Inspired by recent lightweight structures \cite{howard2017mobilenets,ma2018shufflenet,sandler2018mobilenetv2,zhang2018shufflenet}, naive Lite-HRNet \cite{yu2021lite} simply combines the shuffle block in ShuffleNetV2 \cite{ma2018shufflenet} and Small HRNet by replacing residual blocks. 
In this work, separable convolutions are used to replace the normal convolutions, which greatly reduces the computational complexity with little performance loss. 
Based on it, Lite-HRNet \cite{yu2021lite} proposes a more efficient unit, the conditional channel weighting block, to replace the point-wise convolutions in naive Lite-HRNet. 
To achieve the information exchange across channels, point-wise (1×1) convolutions are often used along with separable convolutions, as separable convolutions do not exchange information across channels, and in this case, the point-wise convolutions mainly account for the computational complexity of the model. 
Lite-HRNet applies a channel attention mechanism to serve a similar capability as point-wise convolutions by computing weights conditioned on the input features. 

However, in these methods of human pose estimation, the shapes of the conditional weights at each stage are inconsistent while the scale of the features at each branch remains the same. Especially in the latter stages, high-resolution features are down-sampled to low-resolution features to compute weights, during which high-resolution semantic information may be lost to some extent. In addition, although conditional channel weighting exchanges information across different branches and channels, with the deepening of the network, the parameter numbers of weights in the latter stages increase sharply, causing unnecessary computation costs, which are also inconsistent with the former stages. Furthermore, the global average pooling (GAP) operation are used between the separable convolution and the Squeeze-and-Excitation (SE) operation as the spatial weight computation in Lite-HRNet. The spatial information extraction is not sufficient in this process, as the attention weights are aggregated by a simple averaging method, and the pixel-level pairwise relationship is not taken into consideration, nor is the global context. Also, information exchange across channels only takes place during the SE operation.

To address the above problems, we present a Grouped lightweight High-Resolution Network (Greit-HRNet) for human pose estimation. To extract high-resolution features and to make our method lightweight, we implement a model whose overall architecture is like Lite-HRNet and propose two lightweight units to improve the performance of the model. First, we propose a Grouped Channel Weighting (GCW) method to maintain the consistency of weights among stages while computing weights, in which information exchange across different resolutions and channels takes place as well, and to alleviate the problem of computation redundancy and high-resolution information loss. Then, we introduce a Global Spatial Weighting (GSW) method to exchange information adequately across channels while computing spatial weights. Finally, we apply these two modules to our lightweight high-resolution network for computing weights and extracting features, so that the weights of each branch in the network remain relatively consistent and stable in terms of the scale and the number of parameters in each stage, and fully exchange information across channels during spatial weighting, thus implementing our Greit-HRNet.
\vspace{2\baselineskip}

The main contributions of this work can be summarized as follows:
\begin{itemize}
\item We present a lightweight high-resolution network, Greit-HRNet for human pose estimation, which maintains the stability of weights across stages and strengthens the process of extracting global spatial information in feature maps.
\item We propose a GCW method to keep the scale and the number of parameters of weights relatively consistent across stages and a GSW method to reinforce global information extraction with features exchange across channels, and we adopt these lightweight units in our Greit-HRNet.
\item Our Greit-HRNet achieves the state-of-the-art trade-off between the network performance and complexity on both MS-COCO and MPII human pose estimation datasets.
\end{itemize}

\begin{figure}[tb]
  \centering
  \includegraphics[width=\textwidth]{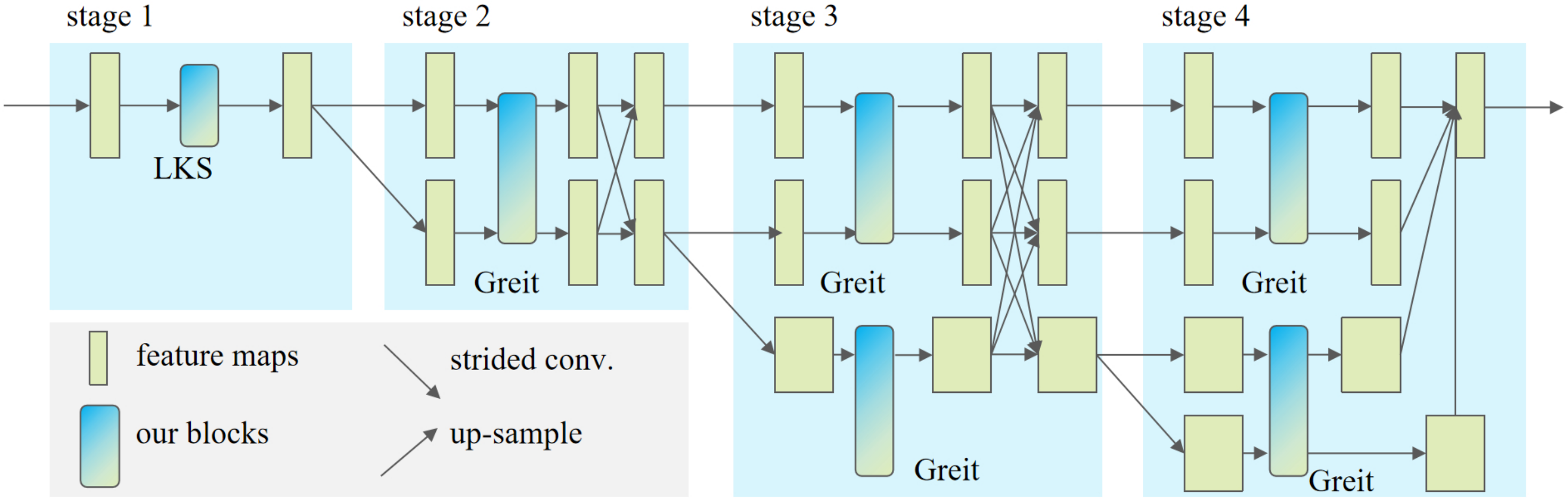}
  \caption{The architecture of Greit-HRNet. The stages go deep in the horizontal direction and the branches expand in the vertical direction.
  }
  \label{fig:fig1}
\end{figure}

\section{Related Work}

\subsubsection{Efficient Methods for Human Pose Estimation.}
For human pose estimation tasks, efficient methods can be adopted to detect key points and estimate poses. Recent works carry out a series of lightweight attempts to achieve a better trade-off between performance and complexity. Given the complexity of 3×3 convolutions, the combination of depth-wise convolutions and point-wise convolutions is widely used in MobileNetV1 \cite{howard2017mobilenets} in place of the normal convolutions to reduce the overall model complexity. ShuffleNet \cite{zhang2018shufflenet} introduces channel shuffling to achieve information exchange across channels to reduce the use of costly point-wise convolutions. Small HRNet \cite{wang2020deep} is proposed with less width and depth and serves as a lightweight backbone network that can be modified for subsequent works, although it has a certain loss of accuracy compared to the original. Inspired by ShuffleNetV2 \cite{ma2018shufflenet} and Small HRNet, Lite-HRNet \cite{yu2021lite} replaces the residual blocks in the Small HRNet with shuffle blocks and further improves the model performance by introducing conditional channel weighting to exchange information across channels and branches.

\subsubsection{Channel and Spatial Attention Mechanisms.}
During convolution operation, both spatial and channel-wise information are perceived by the filter at the same time. E-HRNet \cite{kim2023hrnet} implements the communication among channels by SE operation on the feature maps, and takes the calculated results as the weights, thus achieving a channel attention mechanism. Convolutional Block Attention Module \cite{woo2018cbam,wang2021automated,zhang2022convolutional,farag2022automatic} disentangles the process into two separate dimensions, the channel attention mechanism, and the spatial attention mechanism, to improve the model performance. Non-local Neural Networks \cite{wang2018non} and ELAN \cite{zhang2022efficient} focus on capturing long-range dependencies to extract global spatial information with a spatial attention mechanism. Efficient Multi-Scale Attention Module \cite{ouyang2023efficient} implements a cross-spatial mechanism to capture both short and long-range dependencies.

\subsubsection{Group Methods.}
Grouping input feature maps is an effective way to balance model complexity with model performance. Multi-scale grouped dense network (MSGDN) \cite{li2020multi} groups the input maps and computes the features of different scales separately to obtain satisfactory model performance. Grouped convolutional attention network (GCAnet) \cite{tan2024object} decomposes feature maps into multiple groups with attention modules followed and computed independently to improve the results of feature extraction.

\subsubsection{Large Kernel.}
Recent works find significant advantages to using the large kernel in networks \cite{huang2023large,ding2022scaling}. The separable convolution with the large kernel brings the advantages of a larger receptive field, reducing the number of parameters and the computational complexity, which helps to improve the performance and efficiency of the model. Visual Attention Network (VAN) \cite{guo2023visual} introduces a linear attention mechanism, large kernel attention (LKA), with which depth-wise separable convolutions and dilated convolutions are involved. With linear complexity and extreme simplicity, LKA demonstrated excellent performance.

\section{Greit-HRNet}

\subsubsection{Overall Architecture.}
As shown in Fig.~\ref{fig:fig1}. The overall architecture of Greit-HRNet consists of four stages. The first stage contains the only branch of the high-resolution feature map extracted from the original image, and the next stage has one more branch than the previous stage. At each stage, each branch processes a feature map at a scale different from the other branches. Compared to the previous branch, the newly generated branch below processes a feature map with half the resolution and twice the number of channels, containing more abstract semantic information. At the end of each stage, feature maps of different scales are fused and output as input of multi-scale feature maps for the next stage, with which up-sampling and down-sampling are involved. The final stage outputs only the results of the highest resolution branch for pose estimation. For the fusion method, except for the last stage, the fusion method of other stages is the direct addition. The fusion method adopted in the last stage is to use a convolution operation after each addition, which introduces a few parameters and improves the result.

\subsubsection{Block Arrangement.}
As shown in Table~\ref{tab:tab1}, our blocks have a different number of repetitions at each stage. In each branch of each stage mentioned above (except the first stage), the Greit blocks are repeated several times, as the two Greit blocks form a basic block, and the basic block is repeated several times in each branch. For the rest of this research, we use \textit{a Greit block} to refer to a string of such repeated Greit blocks. For the only branch in the first stage, a Large Kernel Stem (LKS) is used for the initial processing of the feature map of the original image. As shown in Fig.~\ref{fig:fig1}, in other stages, a branch is assigned to a Greit block for processing feature maps, while a Greit block corresponds to one or two branches, if any. The detailed matching relationship is shown in Table~\ref{tab:tab2}. For fair comparisons with Lite-HRNet-18 \cite{yu2021lite} and Lite-HRNet-30 \cite{yu2021lite}, we implement two versions of our network, Greit-HRNet-18 and Greit-HRNet-30 with different network configurations. The detailed network structure is shown in Table~\ref{tab:tab1}.

\begin{figure}[tb]
  \centering
  \begin{subfigure}{0.39\linewidth}
    \includegraphics[width=\linewidth]{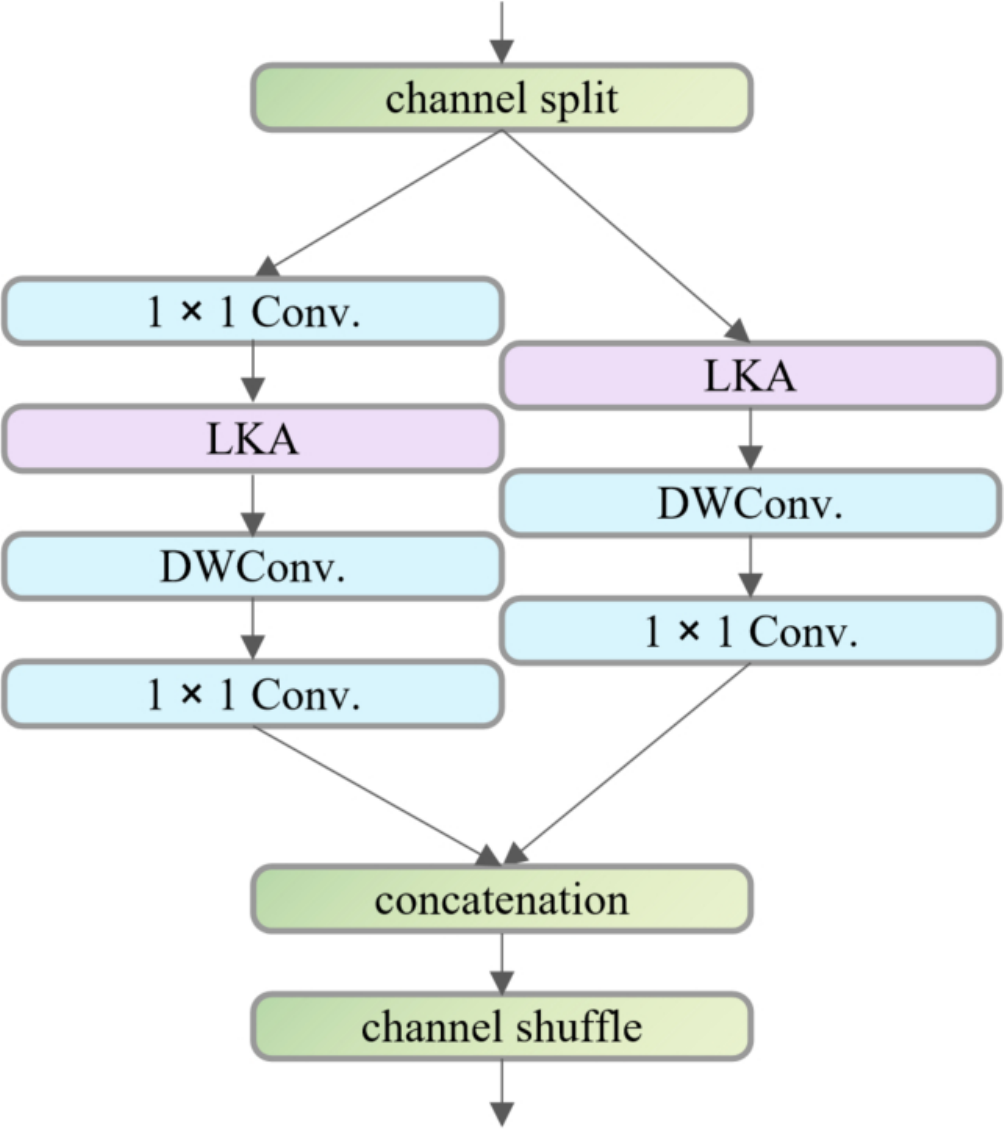}
    \caption{LKS block}
    \label{fig:fig2a}
  \end{subfigure}
  \hfill
  \begin{subfigure}{0.59\linewidth}
    \includegraphics[width=\linewidth]{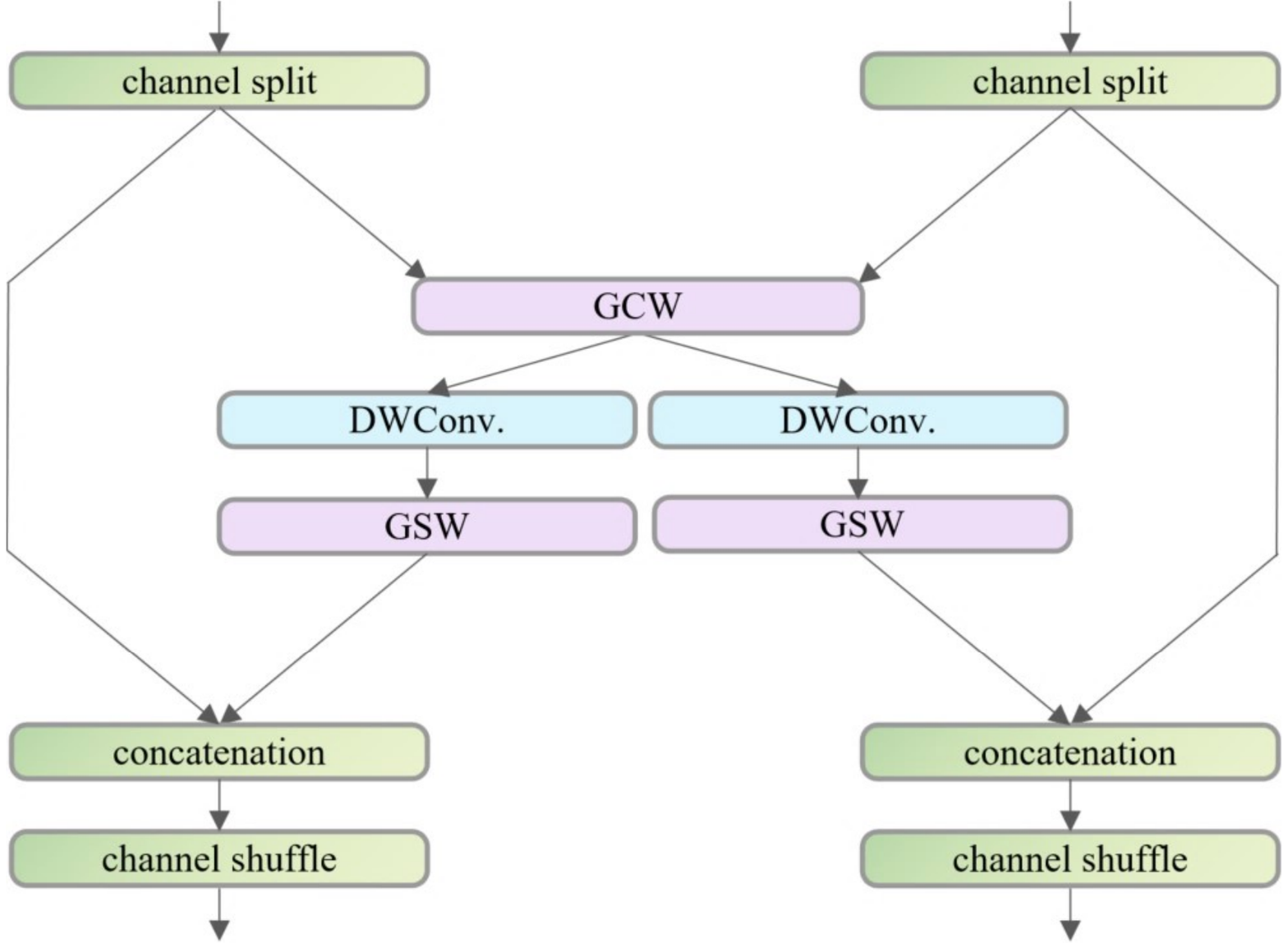}
    \caption{Greit block}
    \label{fig:fig2b}
  \end{subfigure}
  \caption{The LKS block and Greit block. An LKS block applies two LKA modules \cite{guo2023visual} separately. A Greit block corresponds to two branches, if any. The two branches share a GCW module while having a GSW module each. Thus, a Greit block contains a GCW module and two GSW modules (if any).}
  \label{fig:fig2}
\end{figure}

\begin{table}[tb]
  \caption{The network structure. \# denotes the number of. × denotes the repetitions of a certain block as a basic block. Repetition denotes the repetition of the basic block within a stage.}
  \label{tab:tab1}
  \centering
  \begin{tabular}{@{}ccccc@{}}
\toprule
\multirow{2}{*}{Stage} & \multirow{2}{*}{\#Branches} & \multirow{2}{*}{Blocks} & \multicolumn{2}{c}{\#Repetitions} \\
                       &                             &                         & Greit-HRNet-18  & Greit-HRNet-30  \\ \midrule
1                      & 1                           & LKS(×1)                 & 1               & 1               \\
2                      & 2                           & Greit(×2)               & 2               & 3               \\
3                      & 3                           & Greit(×2)               & 4               & 8               \\
4                      & 4                           & Greit(×2)               & 2               & 3               \\ \bottomrule
\end{tabular}
\end{table}

\begin{table}[tb]
  \caption{The detailed matching relationship. There is no branch 4 in stage 3, so the low-resolution group in stage 3 contains only branch 3.}
  \label{tab:tab2}
  \centering
  \begin{tabular}{@{}clclc@{}}
\toprule
GCW module/Greit block           &  & Branch &  & Stage(s) \\ \midrule
\multirow{2}{*}{High-resolution} &  & 1      &  & 2, 3, 4  \\
                                 &  & 2      &  & 2, 3, 4  \\ \midrule
\multirow{2}{*}{Low-resolution}  &  & 3      &  & 3, 4     \\
                                 &  & 4      &  & 4        \\ \bottomrule
\end{tabular}
  
\end{table}

\subsubsection{Block Structure.}
As shown in Fig.~\ref{fig:fig2}, the LKS and Greit block follow an architecture like ShuffleNetV2 \cite{ma2018shufflenet}. The feature map is split into two subdivisions along the channel dimension. After processing the two subdivisions to different degrees, they are concatenated with a channel shuffle operation behind them. In LKS, shown in Fig.~\ref{fig:fig2a}, one subdivision first applies an LKA module \cite{guo2023visual}, a depth-wise separable convolution, and a point-wise convolution, while the other subdivision expands the feature map along the channel dimension first and restores it later with a point-wise convolution. In Greit block, shown in Fig.~\ref{fig:fig2b}, the processing of the feature map inside a subdivision is roughly similar to the Convolutional Block Attention Module. A GCW module is first applied to process the feature map, which is a channel attention mechanism, followed by a local sampling with a depth-wise separable convolution. A GSW module then presents a spatial attention mechanism. In addition, two adjacent branches, if any, share a GCW module. The detailed matching relationship is shown in Table~\ref{tab:tab2}.

\subsection{Grouped Channel Weighting}

\subsubsection{Conditional Channel Weighting.}
Conditional channel weighting computes weights conditioned on the input. For the original feature tensor $X$, we compute weights $W$ after processing $P$ on $X$, and obtain the result $Y$ after performing element-wise multiplication operation between weights $W$ and $X$. This process can be written as,
\begin{align}
  Y & = W \odot X, \; \label{eq:eq1}\\
W & = P(X), \label{eq:eq2}
\end{align}
where $\odot$ denotes the element-wise multiplication operator. 

To introduce the channel attention mechanism, the SE operation is applied as P and it improves the efficiency and expression ability during feature extraction. The process then can be rewritten as,
\begin{equation}
  Y=sigmoid(Conv_1(ReLU(Conv_2(X)))) \odot X,
  \label{eq:eq3}
\end{equation}
where $sigmoid()$ and $ReLU()$ denote activation functions, while $Conv_1()$ and $Conv_2()$ denote point-wise convolutions.

\begin{figure}[tb]
  \centering
  \includegraphics[width=0.6\textwidth]{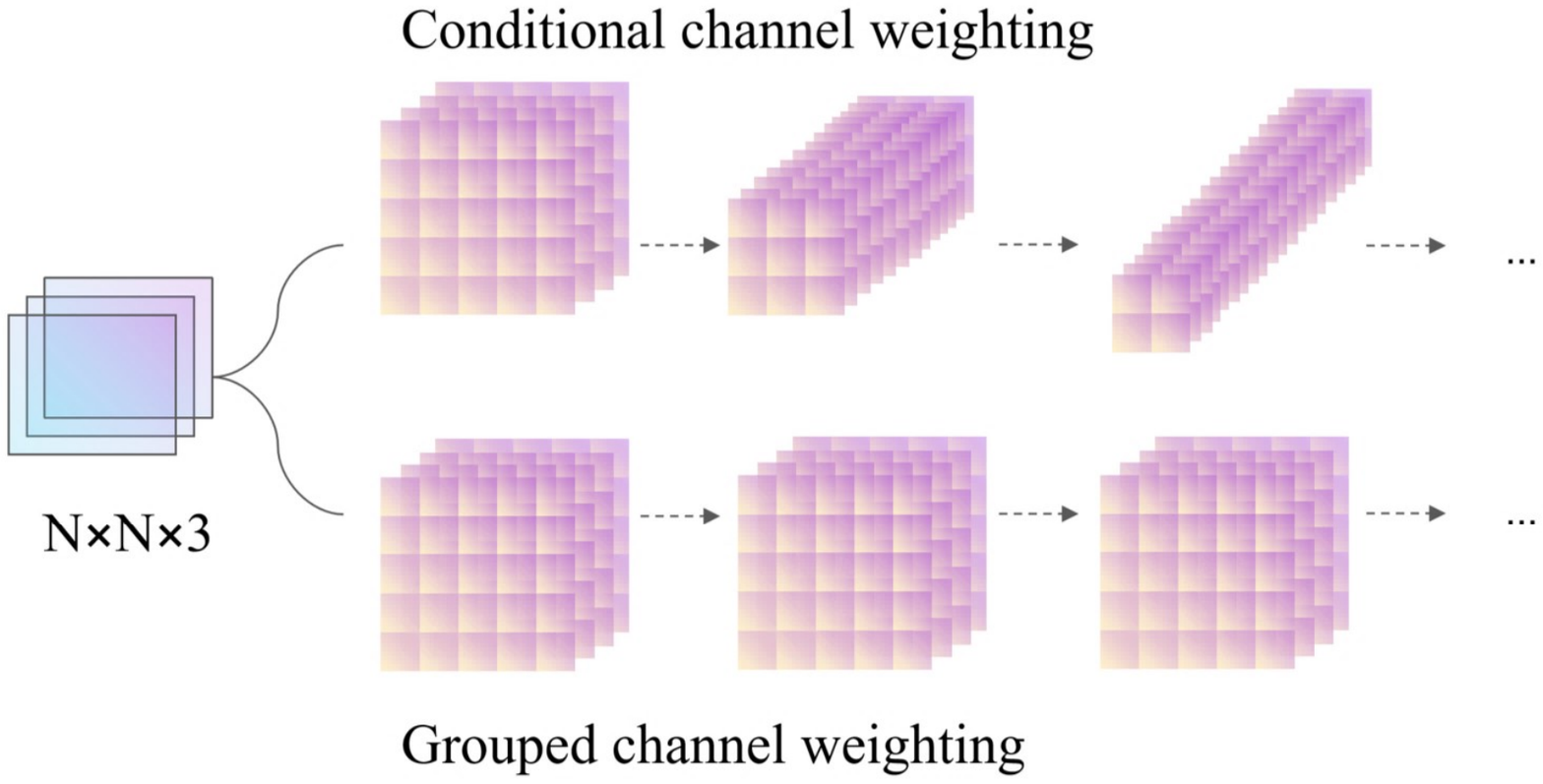}
  \caption{The transition of feature maps with the deepening of stages. For Conditional channel weighting, the number of channels increases rapidly, while the number for our Grouped channel weighting remains stable.
  }
  \label{fig:fig3}
\end{figure}

In Lite-HRNet \cite{yu2021lite}, the $X$ in Equation~\ref{eq:eq1}, Equation~\ref{eq:eq2}, and Equation~\ref{eq:eq3} is computed by the following two steps. First, reshape the tensors $X_1, X_2, \dots, X_{s-1}$ into the shape of $X_s$ through adaptive average pooling (AAP) operation to get the results, $X_1', X_2', \dots, X_s'$, where $X_s$ denotes the tensor of the lowest resolution and $X_i'$ denotes $X_i$ after adaptive average pooling. Second, the results, $X_1', X_2', \dots, X_s'$ are concatenated together along the channel dimension to get $Z$. Then $Z$ is used as the $X$ in Equation~\ref{eq:eq1}, Equation~\ref{eq:eq2}, and Equation~\ref{eq:eq3}. The advantage is that information can be exchanged among resolutions. 
However, in the above process, as the stage deepens, the number of channels $C_z$ of tensor $Z$ grows rapidly, as shown in Fig.~\ref{fig:fig3}. Given that the number of parameters inside $Conv_1$ and $Conv_2$ in Equation~\ref{eq:eq3} is quadratic with $C_z$, this processing results in a rapid increase in the overall model complexity. This is inconsistent with the former stages. In addition, a large amplitude of resolution change leads to the loss of high-resolution information at the latter stages, which is also discordant.

\subsubsection{Group Methods.}
To address these problems, we propose to use the efficient group method to replace the method mentioned above. Firstly, all branches in all stages are grouped according to Table~\ref{tab:tab2}. Specifically, these branches are divided into two groups: the high-resolution group and the low-resolution group. The high-resolution group contains branch 1 and branch 2, while the low-resolution group contains branch 3 and branch 4. Conditional channel weighting operations only take place within a single group with fewer parameters. The method can be written as,
\begin{align}
  Y_h & = sigmoid(Conv_{h1}(ReLU(Conv_{h2}(X_h)))) \odot X_h, \; \\
Y_l & = sigmoid(Conv_{l1}(ReLU(Conv_{l2}(X_l)))) \odot X_l,
\end{align}
where $h$ denotes high-resolution groups, while $l$ denotes low-resolution groups. 

The advantage is that high-resolution features can be maintained during the pooling operation as the stage deepens. In addition, each branch in the same group contains the features of all the scales in the previous fusion layer. Concatenating them and performing SE operations within the group achieves the information exchange among all resolutions.

\begin{figure}[tb]
  \centering
  \begin{subfigure}{0.49\linewidth}
    \includegraphics[width=\linewidth]{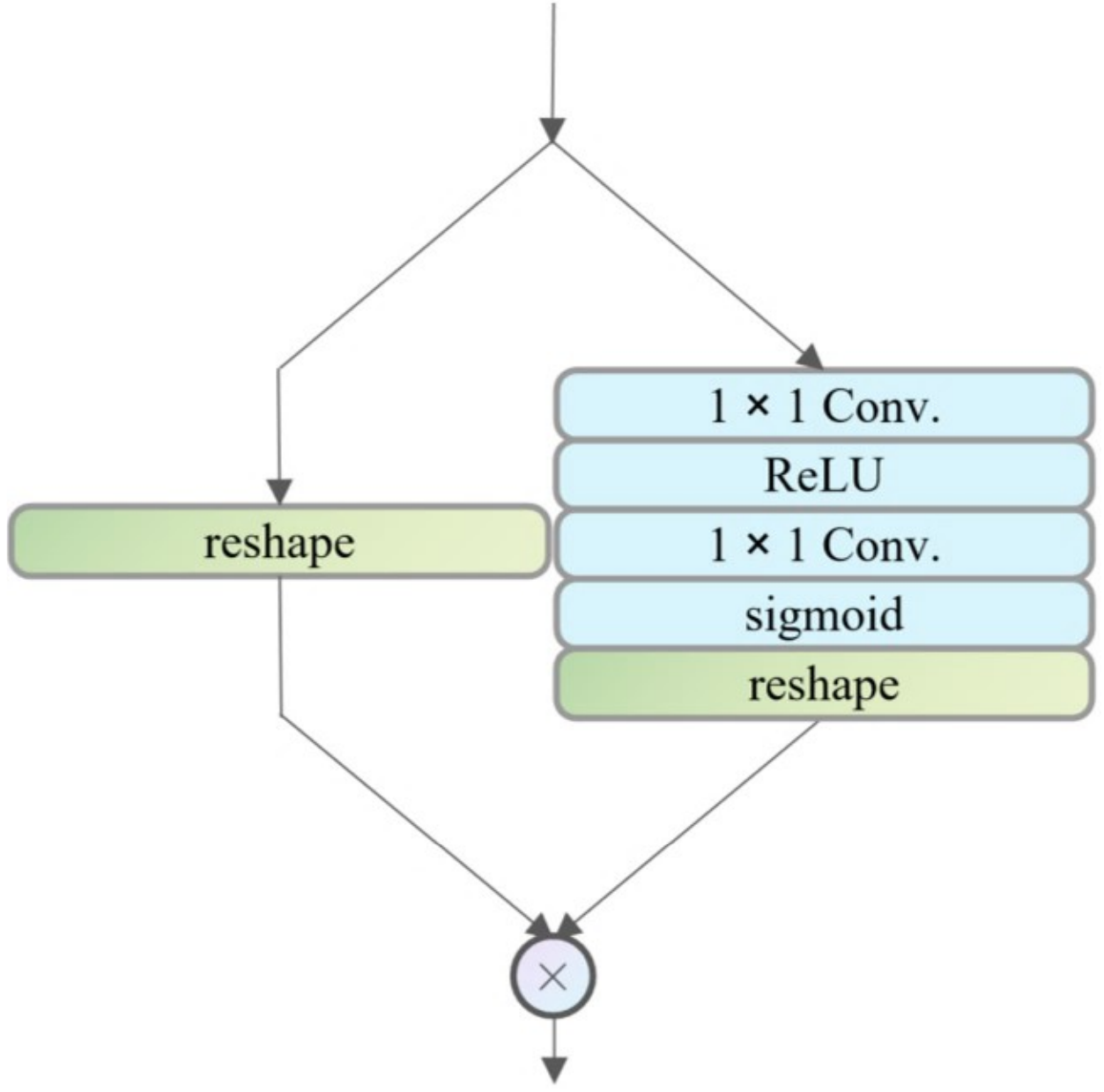}
    \caption{Information extraction}
    \label{fig:fig4a}
  \end{subfigure}
  \hfill
  \begin{subfigure}{0.49\linewidth}
    \includegraphics[width=\linewidth]{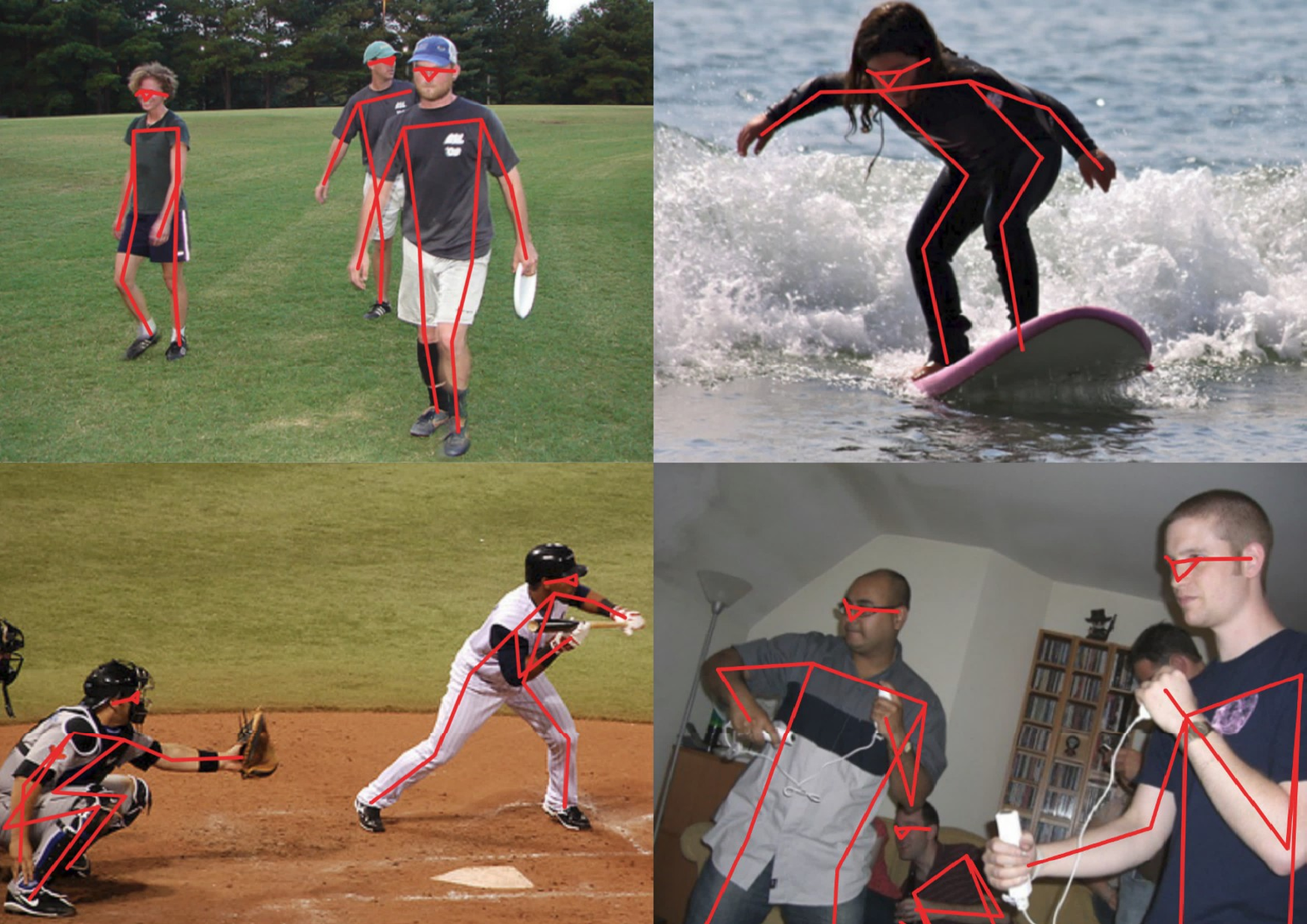}
    \caption{Example qualitative results}
    \label{fig:fig4b}
  \end{subfigure}
  \caption{Global spatial information extraction and example qualitative results. $\otimes$ denotes matrix multiplication. The first convolution sets the number of channels in the feature graph to a hyper-parameter $C'$, and the second convolution reduces this number to 1.}
  \label{fig:fig4}
\end{figure}
\subsection{Global Spatial Weighting}

\subsubsection{Spatial Attention Weighting.}
To extract spatial features, global average pooling (GAP) is introduced to compute the spatial weights. The length of height and width dimensions of the input feature map $X$ is changed to 1 by GAP operation. Then it goes through two convolution layers and performs element-wise multiplication operations with $X$ as spatial attention weights. This process can be written as,
\begin{equation}
  Y = sigmoid(Conv_1(ReLU(Conv_2(GAP(X))))) \odot X,
\end{equation}
where $GAP()$ denotes a GAP operation.

\subsubsection{Global Spatial Information.}
However, in the method mentioned above, the extraction of global context by GAP operation causes spatial information loss. Especially when the feature map is of high-resolution, the addition and average results cannot well represent the global context, as the GAP operation fails to pay attention to the pixel-level pairwise relationship and does not show preference and emphasis within the global context. Thus, we present a method to solve this problem, as shown in Fig.~\ref{fig:fig4a}. For the tensor $X$ with scale C×H×W, where C, H, and W denote the number of channels, height, and width respectively, we first reshape it into the scale 1×C×HW to get $X'$. Next, we perform two convolution operations on the original tensor $X$ to reduce its dimension of the channel dimension, and then reshape to get a tensor $Z$ with a scale of 1×HW×1. Finally, matrix multiplication of $X'$ and $Z$ provides global spatial weights $Y$. This process can be written as,
\begin{align}
  Y & = X' \cdot Z, \; \\
X' & = Re(X), \\
Z & = Re(sigmoid(Conv_1(ReLU(Conv_2(X)))),
\end{align}
where $\cdot$ denotes matrix multiplication and $Re$ denotes reshape operation. 

The $Y$ above is a tensor of a scale of 1×C×1, and after reshaping and a SE operation on it, we get global spatial weights, which are used to perform element-wise multiplication with the original feature map $X$.
In the whole process mentioned above, the pixel-level pairwise relationship has been well noted in matrix multiplication, and the information exchange among channels has been further promoted. In addition, the key point within the global context has also been focused on and emphasized instead of simple addition and average operations without preference.

\subsection{Large Kernel Stem}

\subsubsection{Large Kernel Attention.}
In the Large kernel attention (LKA) method \cite{guo2023visual}, the feature map $X$ is first extracted by the depth-wise convolution and the depth-wise dilated convolution with the large kernel, then by a point-wise convolution to get the attention weights $Z$. Finally, element-wise multiplication is performed on $Z$ and $X$ to get the result $Y$. The LKA method can be written as, 
\begin{align}
  Y & =Z \odot X, \; \\
Z & = Conv(DWDConv(DWConv(X))),
\end{align}
where $DWConv()$ denotes a depth-wise convolution and $DWDConv()$ denotes a depth-wise dilated convolution.

\subsubsection{Simple yet Efficient.}
In this process, the large kernel, dilated, and depth-wise convolutions can greatly expand the receptive field and obtain rich semantic information in the initial processing of the feature map of the input image. It provides effective assistance for the subsequent operation, thus improving the performance of the overall model. In addition, despite the use of the large kernel, the overall complexity of the model does not rise, and the increase in computation is negligible, due to the lightweight design of depth-wise convolutions and the complexity linearly related to the number of channels.

\begin{table}[tb]
  \caption{Comparisons of results on the COCO val2017 set. If Pre is Y, then the backbone is pretrained on the ImageNet classification task or MS-COCO train set, otherwise, the value is N. If Aug is Y, then data augmentations are involved during training, otherwise, the value is N. Bold indicates the best result, and underline indicates the highest score with the lowest \#Params or GFLOPs.}
  \label{tab:tab3}
  \centering
  \resizebox{\textwidth}{!}{%
\begin{tabular}{@{}llcccccccccclc@{}}
\toprule
Model & Backbone & Pre & Aug & Input size & \#Params(M) & GFLOPs & AP & AP\textsuperscript{50} & AP\textsuperscript{75} & AP\textsuperscript{M} & \multicolumn{2}{c}{AP\textsuperscript{L}} & AR \\ \midrule
\multicolumn{14}{c}{Large networks} \\ \midrule
SimpleBaseline \cite{xiao2018simple}& ResNet-50 & Y & Y & 256×192 & 34.0 & 8.9 & 70.4 & 88.6 & 78.3 & 67.1 & \multicolumn{2}{c}{77.2} & 76.3 \\
HRNet \cite{sun2019deep}& HRNet-W32 & N & Y & 256×192 & 28.5 & 7.1 & 73.4 & 89.5 & 80.7 & 70.2 & \multicolumn{2}{c}{80.1} & 78.9 \\
 HigherHRNet-W48 \cite{cheng2020higherhrnet}& HRNet-W48& N& Y& 640×640& 63.8& 154.3& 69.9& /& /& 65.4& \multicolumn{2}{c}{76.4}&/\\
 HRFormer-B \cite{yuan2021hrformer}& HRFormer-B& Y& Y& 256×192& 43.2& 12.2& 75.6& 90.8& 82.8& 71.7& \multicolumn{2}{c}{82.6}&80.8\\
 ViTPose-B \cite{xu2022vitpose}& ViT-B& Y& N& 256×192& 86& /& 75.8& /& /& /& \multicolumn{2}{c}{/}&81.1\\ \midrule
\multicolumn{14}{c}{Small networks} \\ \midrule
MobileNetV2 \cite{sandler2018mobilenetv2}& MobileNetV2 & N & N & 256×192 & 9.6 & 1.4 & 64.6 & 87.4 & 72.3 & 61.1 & \multicolumn{2}{c}{71.2} & 70.7 \\
ShuffleNetV2 \cite{ma2018shufflenet}& ShuffleNetV2 & N & N & 256×192 & 7.6 & 1.2 & 59.9 & 85.4 & 66.3 & 56.6 & \multicolumn{2}{c}{66.2} & 66.4 \\
Small HRNet \cite{wang2020deep}& HRNet-W18 & N & Y & 256×192 & 1.3 & 0.5 & 55.2 & 83.7 & 62.4 & 52.3 & \multicolumn{2}{c}{61.0} & 62.1 \\
\multirow{2}{*}{Lite-HRNet \cite{yu2021lite}} & Lite-HRNet-18 & N & Y & 256×192 & \textbf{1.1} & \textbf{0.2} & 64.8 & 86.7 & 73.0 & 62.1 & \multicolumn{2}{c}{70.5} & 71.2 \\
 & Lite-HRNet-30 & N & Y & 256×192 & 1.8 & 0.3 & 67.2 & 88.0 & 75.0 & 64.3 & \multicolumn{2}{c}{73.1} & 73.3 \\
 \multirow{2}{*}{X-HRNet \cite{zhou2022x}}& X-HRNet-18& N& Y& 256×192& 1.3& \textbf{0.2}& 65.1& 86.7& 72.7& 62.3& \multicolumn{2}{c}{70.9}& 71.2\\
 & X-HRNet-30& N& Y& 256×192& 2.1& 0.3& 67.4& 87.5& 75.4& 64.5& \multicolumn{2}{c}{73.3}& 73.5\\
\multirow{2}{*}{\textbf{Greit-HRNet}} & \textbf{Greit-HRNet-18} & N & Y & 256×192 & \textbf{1.1} & \textbf{0.2} & {\ul 65.8} & {\ul 87.1} & {\ul 73.8} & {\ul 63.2} & \multicolumn{2}{c}{{\ul 71.3}} & {\ul 71.8} \\
 & \textbf{Greit-HRNet-30} & N & Y & 256×192 & 1.8 & 0.4 & 68.2 & 88.1 & 76.0 & 65.4 & \multicolumn{2}{c}{73.8} & 74.0 \\ \midrule
MobileNetV2 \cite{sandler2018mobilenetv2}& MobileNetV2 & N & N & 384×288 & 9.6 & 3.3 & 67.3 & 87.9 & 74.3 & 62.8 & \multicolumn{2}{c}{74.7} & 72.9 \\
ShuffleNetV2 \cite{ma2018shufflenet}& ShuffleNetV2 & N & N & 384×288 & 7.6 & 2.8 & 63.6 & 86.5 & 70.5 & 59.5 & \multicolumn{2}{c}{70.7} & 69.7 \\
Small HRNet \cite{wang2020deep}& HRNet-W18 & N & Y & 384×288 & 1.3 & 1.2 & 56.0 & 83.8 & 63.0 & 52.4 & \multicolumn{2}{c}{62.6} & 62.6 \\
\multirow{2}{*}{Lite-HRNet \cite{yu2021lite}} & Lite-HRNet-18 & N & Y & 384×288 & \textbf{1.1} & 0.4 & 67.6 & 87.8 & 75.0 & 64.5 & \multicolumn{2}{c}{73.7} & 73.7 \\
 & Lite-HRNet-30 & N & Y & 384×288 & 1.8 & 0.7 & 70.4 & 88.7 & 77.7 & 67.5 & \multicolumn{2}{c}{76.3} & 76.2 \\
 EfficientHRNet-H$_0$ \cite{neff2021efficienthrnet}& EffcientNet& N& Y& 512×512& 23.3& 25.6& 64.8& /& /& /& \multicolumn{2}{c}{/}& / \\
 LitePose-S \cite{wang2022lite}& MobileNetV2& Y& Y& 448×448& 2.7& /& 56.8& /& /& /& \multicolumn{2}{c}{/}& / \\
 \multirow{2}{*}{X-HRNet \cite{zhou2022x}}& X-HRNet-18& N& Y& 384×288& 1.3& 0.4& 67.9& 87.6& 75.5& 64.7& \multicolumn{2}{c}{73.9}& 73.6\\
 & X-HRNet-30& N& Y& 384×288& 2.1& 0.6& 70.6& 88.9& 77.7& 67.6& \multicolumn{2}{c}{76.5}& 76.1\\
\multirow{2}{*}{\textbf{Greit-HRNet}} & \textbf{Greit-HRNet-18} & N & Y & 384×288 & \textbf{1.1} & 0.6 & {\ul 68.8} & {\ul 87.9} & {\ul 75.9} & {\ul 65.4} & \multicolumn{2}{c}{{\ul 75.3}} & {\ul 75.1} \\
 & \textbf{Greit-HRNet-30} & N & Y & 384×288 & 1.8 & 0.9 & \textbf{71.4} & \textbf{88.8} & \textbf{78.0} & \textbf{68.1} & \multicolumn{2}{c}{\textbf{77.5}} & \textbf{77.3} \\ \bottomrule
\end{tabular}%
}
  
\end{table}

\begin{table}[tb]
  \caption{Comparisons of results on the COCO test-dev2017 set. Bold indicates the best result and underline indicates the highest score with the lowest \#Params or GFLOPs.}
  \label{tab:tab4}
  \centering
\resizebox{\textwidth}{!}{%
\begin{tabular}{@{}llccccccccclcc@{}}
\toprule
Model & Backbone & Pre & Aug & Input size & \#Params(M) & GFLOPs & AP & AP\textsuperscript{50} & AP\textsuperscript{75} & \multicolumn{2}{c}{AP\textsuperscript{M}} & AP\textsuperscript{L} & AR \\ \midrule
\multicolumn{14}{c}{Large networks} \\ \midrule
SimpleBaseline \cite{xiao2018simple}& ResNet-50 & Y & Y & 256×192 & 34.0 & 8.9 & 70.0 & 90.9 & 77.9 & \multicolumn{2}{c}{66.8} & 75.8 & 75.6 \\
HRNet \cite{sun2019deep}& HRNet-W32 & N & Y & 384×288 & 28.5 & 16.0 & 74.9 & 92.5 & 82.8 & \multicolumn{2}{c}{71.3} & 80.9 & 80.1 \\ 
HigherHRNet-W48 \cite{cheng2020higherhrnet}& HRNet-W48& N& Y& 640×640& 63.8& 154.3& 68.4& 88.2& 75.1& 64.4& \multicolumn{2}{c}{74.2}& / \\
HRFormer-B \cite{yuan2021hrformer}& HRFormer-B& Y& Y& 384×288& 43.2& 26.8& 76.2& 92.7& 83.8& 72.5& \multicolumn{2}{c}{82.3}& 81.2 \\
\midrule
\multicolumn{14}{c}{Small networks} \\ \midrule
MobileNetV2 \cite{sandler2018mobilenetv2}& MobileNetV2 & N & N & 384×288 & 9.6 & 3.3 & 66.8 & 90.0 & 74.0 & \multicolumn{2}{c}{62.6} & 73.3 & 72.3 \\
ShuffleNetV2 \cite{ma2018shufflenet}& ShuffleNetV2 & N & N & 384×288 & 7.6 & 2.8 & 62.9 & 88.5 & 69.4 & \multicolumn{2}{c}{58.9} & 69.3 & 68.9 \\
Small HRNet \cite{wang2020deep}& HRNet-W18 & N & Y & 384×288 & 1.3 & 1.2 & 55.2 & 85.8 & 61.4 & \multicolumn{2}{c}{51.7} & 61.2 & 61.5 \\
\multirow{2}{*}{Lite-HRNet \cite{yu2021lite}} & Lite-HRNet-18 & N & Y & 384×288 & \textbf{1.1} & \textbf{0.4} & 66.9 & 89.4 & 74.4 & \multicolumn{2}{c}{64.0} & 72.2 & 72.6 \\
 & Lite-HRNet-30 & N & Y & 384×288 & 1.8 & 0.7 & 69.7 & 90.7 & 77.5 & \multicolumn{2}{c}{66.9} & 75.0 & 75.4 \\
EfficientHRNet-H$_0$ \cite{neff2021efficienthrnet}& EffcientNet& N& Y& 512×512& 23.3& 25.6& 64.0& /& /& /& \multicolumn{2}{c}{/}& / \\
LitePose-S \cite{wang2022lite}& MobileNetV2& Y& Y& 448×448& 2.7& 25.6& 56.7& /& /& /& \multicolumn{2}{c}{/}& / \\
 \multirow{2}{*}{X-HRNet \cite{zhou2022x}}& X-HRNet-18& N& Y& 384×288& 1.3& 0.4& 67.3& 89.8& 74.8& /& \multicolumn{2}{c}{/}& 73.0\\
 & X-HRNet-30& N& Y& 384×288& 2.1& 0.6& 70.0& 90.6& 77.7& /& \multicolumn{2}{c}{/}& 75.5\\
\multirow{2}{*}{\textbf{Greit-HRNet}} & \textbf{Greit-HRNet-18} & N & Y & 384×288 & \textbf{1.1} & 0.6 & {\ul 68.2} & {\ul 89.7} & {\ul 75.6} & \multicolumn{2}{c}{{\ul 65.0}} & {\ul 73.5} & {\ul 74.1} \\
 & \textbf{Greit-HRNet-30} & N & Y & 384×288 & 1.8 & 0.9 & \textbf{70.5} & \textbf{90.6} & \textbf{78.1} & \multicolumn{2}{c}{\textbf{67.2}} & \textbf{76.0} & \textbf{76.1} \\ \bottomrule
\end{tabular}%
}
\end{table}

\begin{table}[tb]
  \caption{Comparisons of results on the MPII val set. Bold indicates the best result and underline indicates the second-best result.}
  \label{tab:tab5}
  \centering
\begin{tabular}{@{}lccc@{}}
\toprule
Model & \#Params(M) & GFLOPs & PCKh \\ \midrule
MobileNetV2 \cite{sandler2018mobilenetv2}& 9.6 & 1.9 & 85.4 \\
ShuffleNetV2 \cite{ma2018shufflenet}& 7.6 & 1.7 & 82.8 \\
Small HRNet \cite{wang2020deep}& {\ul 1.3} & 0.7 & 80.2 \\
Lite-HRNet-18 \cite{yu2021lite}& \textbf{1.1} & \textbf{0.2} & 86.1 \\
Lite-HRNet-30 \cite{yu2021lite}& 1.8 & 0.4 & {\ul 87.0} \\
Greit-HRNet-18 & \textbf{1.1} & {\ul 0.3} & 86.8 \\
Greit-HRNet-30 & 1.8 & 0.5 & \textbf{87.4} \\ \bottomrule
\end{tabular}
\end{table}

\section{Experiments}
To evaluate our approach, we report the performance of our model compared to start-of-arts on two human pose estimation datasets, the MS-COCO \cite{lin2014microsoft} and the MPII \cite{andriluka20142d}. In addition, comprehensive ablation studies on the MS-COCO dataset are performed to demonstrate the effectiveness of our modules.

\subsection{Configurations}

\subsubsection{Datasets.}
The images of the MS-COCO dataset contain over 250K person instances with annotations of 17 key points each. The training of our Greit-HRNet is on the train2017 set with 57K images and 150K person instances included. The evaluation is on the val2017 set and test-dev2017 set with 5K images and 20K images included, respectively. In addition, the performance of our models is also studied on the MPII Human Pose dataset with around 25K images and 40K person instances included, of which thirty percent are used for testing and the rest for training.

\subsubsection{Training.}
Our Greit-HRNet is trained on 8 GeForce RTX 4090 GPUs with a batch size of 32 per GPU. Adam optimizer is adopted with an initial learning rate of $2e^{-3}$ to update parameters, while this number gradually decreases and stabilizes at $2e^{-5}$ at the final epochs of more than 200. For data processing, human detection boxes are extended to a fixed aspect ratio of 4:3, which are used to crop the images, and the images are resized to 256×192 or 384×288 for the MS-COCO dataset, and 256×256 for the MPII dataset. For data augmentations, random rotations with the factor of 30 ([-30°, 30°]), random scales with the factor of 0.25 ([0.75, 1.25]), and random flipping are engaged for both the MS-COCO and the MPII datasets. In addition, extra half-body transformations are performed for the MS-COCO dataset.

\subsubsection{Testing.}
For the MS-COCO dataset, the two-stage top-down paradigm is adopted for testing our models. The person detection boxes are first produced and the person key points are then predicted with person boxes predicted by the person detectors provided by SimpleBaseline \cite{xiao2018simple}. For the MPII dataset, the standard testing strategy is adopted with the provided person boxes. The heat maps are estimated via 2D Gaussian, followed by an average operation between the original and altered images. A quarter offset is performed to adjust the key-point location, whose direction is from the highest to the second-highest response.

\subsubsection{Evaluation.}
Average Precision (AP) and Average Recall (AR) scores based on Object Key-point Similarity (OKS) are adopted to evaluate the performance of our models on the MS-COCO dataset. The accuracy of the head-normalized Probability of Correct Key-point (PCKh) score is evaluated for the MPII dataset.

\subsection{Results}

\subsubsection{Results on the COCO val2017 Set.}
As shown in Table~\ref{tab:tab3}, we report the performance of our methods Greit-HRNet-18 and Greit-HRNet-30 in comparison with other state-of-the-art methods on the COCO val2017 Set. For lightweight networks, we train Greit-HRNet with input sizes 384 × 288 to achieve the highest AP score of 71.4, showing a better performance than other methods. For MobileNetV2 \cite{sandler2018mobilenetv2}, ShuffleNetV2 \cite{ma2018shufflenet}, and EfficientHRNet-H$_0$ \cite{neff2021efficienthrnet}, our model achieves better accuracy while reducing the number of model parameters by about 80$\sim$95\% of them. For Small HRNet \cite{wang2020deep} and LitePose-S \cite{wang2022lite}, our model has fewer parameters and less computational complexity but is more accurate. In addition, our Greit-HRNet has a comparable amount of parameters and computational cost as Lite-HRNet \cite{yu2021lite} and X-HRNet \cite{zhou2022x} but has higher results than them in AP, achieving a better trade-off between model complexity and performance. For large networks, our Greit-HRNet achieves a comparable accuracy with significantly smaller model complexity compared to them. Fig.~\ref{fig:fig4b} shows the visual results on COCO from Greit-HRNet-30.

\subsubsection{Results on the COCO test-dev2017 Set.}
As shown in Table~\ref{tab:tab4}, we report the results on the COCO test-dev2017 Set of our Greit-HRNet and other state-of-the-art methods and demonstrate the superior efficiency of our method. Compared with Lite-HRNet, our Greit-HRNet-18 improves the AP and AR by 1.3\% and 1.5\% respectively, while the model and computational complexity are almost unchanged. In addition, despite a small loss of accuracy, compared with large networks, the computational cost and our model complexity are reduced by about 95\%.

\subsubsection{Results on the MPII val Set.}
As shown in Table~\ref{tab:tab5}, we report the results on the MPII val Set to show the superior performance of our Greit-HRNet on this dataset compared to other lightweight models. Compared to Lite-HRNet-18 and Lite-HRNet-30, our models improve by 0.7 and 0.4 points on the PCKh@0.5 score, respectively, with the same amount of model parameters and computational complexity. Compared to MobileNetV2 and ShuffleNetV2, our models have considerably fewer parameters and computational loss and achieve an accuracy of more than 2.0~4.0 points higher than them. Our Greit-HRNet-30 achieves the highest score of 87.4 PCKh@0.5, which shows the superiority of our model in optimizing the trade-off between model capacity and complexity.

Due to the introduction of the efficient Greit and LKS blocks, our Greit-HRNet maintains the consistency of weights across stages and improves the process of extracting global spatial information in feature maps, achieving the state-of-the-art trade-off between the network capacity and complexity on both MS-COCO and MPII human pose estimation datasets.

\begin{table}[tb]
  \caption{Ablation studies on the COCO val2017.}
  \label{tab:tab6}
  \centering
\begin{tabular}{@{}llll@{}}
\toprule
Model & \#Param(M) & GFLOPs & AP \\ \midrule
Lite-HRNet-18 \cite{yu2021lite}& 1.13 & 0.2 & 64.8 \\
with GCW & 1.16 & 0.2 & 65.2 \\
with GCW\&GSW & 1.15 & 0.2 & 65.6 \\
with GSW & 1.12 & 0.2 & 65.2 \\
with LKS & 1.13 & 0.2 & 65.0 \\
with LKS\&GCW & 1.12 & 0.2 & 65.4 \\
with LKS\&GSW & 1.16 & 0.2 & 65.4 \\
Greit-HRNet-18 & 1.15 & 0.2 & 65.8 \\ \bottomrule
\end{tabular}
\end{table}

\subsection{Ablation Study}
We conduct ablation experiments on the three modules, GCW, GSW, and LKS, and report their respective performances on the COCO val2017 set with input size 256×192, as shown in Table~\ref{tab:tab6}. To demonstrate the superior efficiency of our method, we replace conditional channel weighting modules in Lite-HRNet \cite{yu2021lite} with two stacked GCW modules to have similar model complexity and computational cost as Lite-HRNet. The overall architecture shown in Table~\ref{tab:tab1} is used for the subsequent ablation comparison experiments.

\subsubsection{Effectiveness of GCW.}
We take Lite-HRNet as the baseline, on which we first introduce our GCW method to replace conditional channel weighting. Compared with the original network, our GCW modules introduce only a small amount of parameters and computational cost but improve the result from 64.8\% to 65.2\%. Compared with networks using LKS, LKS\&GSW, and GSW methods, the performance of the networks with GCW modules introduced also increases by about 0.4\%.

\subsubsection{Effectiveness of GSW.}
Replacing the spatial weight computation modules in the baseline with GSW modules, the number of parameters in the network decreases, and the overall performance of the network is improved. Compared with the networks using LKS, LKS\&GCW, and GCW methods, the results increase from 65.0\%, 65.4\%, and 65.2\% to 65.4\%, 65.8\%, and 65.6\% respectively.

\subsubsection{Effectiveness of LKS.}
Compared with Lite-HRNet, after the introduction of LKS, the model complexity and computational complexity cost are not significantly improved, and the overall performance of the model increases. For networks that use GCW, GSW, and GCW\&GSW methods, the introduction of the LKS method increases their original results by 0.2\%, demonstrating the efficiency of LKS modules.

\section{Conclusion}
In this work, we present a lightweight and efficient network, Greit-HRNet, for human pose estimation. To solve the inconsistency with the deepening of the network and improve the spatial information extraction in the past works, we propose a Greit block, in which GCW and GSW methods are introduced to maintain the stability of weights and extract spatial information effectively in the network. It provides a solution to solve the problems and improve the efficiency of the model. In addition, the introduction of LKS blocks is a simple yet efficient way to improve the overall performance of the model. Our network has achieved good results on both the MS-COCO and the MPII datasets, providing a novel method to optimize lightweight high-resolution networks.

\begin{credits}
\subsubsection{\ackname} 
 This study was funded by Professor Yingxia Yu. We would like to express our sincere gratitude to Professor Yanxia Wang for her invaluable guidance throughout this research. We also extend our heartfelt thanks to Shuyao Shang for his assistance and support.

\subsubsection{\discintname}
The authors have no competing interests to declare that are relevant to the content of this article.
\end{credits}
%
%
%
\bibliographystyle{splncs04}
\bibliography{main}

\begin{thebibliography}{10}
\providecommand{\url}[1]{\texttt{#1}}
\providecommand{\urlprefix}{URL }
\providecommand{\doi}[1]{https://doi.org/#1}

\bibitem{andriluka20142d}
Andriluka, M., Pishchulin, L., Gehler, P., Schiele, B.: 2d human pose estimation: New benchmark and state of the art analysis. In: Proceedings of the IEEE Conference on computer Vision and Pattern Recognition. pp. 3686--3693 (2014)

\bibitem{cheng2020higherhrnet}
Cheng, B., Xiao, B., Wang, J., Shi, H., Huang, T.S., Zhang, L.: Higherhrnet: Scale-aware representation learning for bottom-up human pose estimation. In: Proceedings of the IEEE/CVF conference on computer vision and pattern recognition. pp. 5386--5395 (2020)

\bibitem{ding2022scaling}
Ding, X., Zhang, X., Han, J., Ding, G.: Scaling up your kernels to 31x31: Revisiting large kernel design in cnns. In: Proceedings of the IEEE/CVF conference on computer vision and pattern recognition. pp. 11963--11975 (2022)

\bibitem{farag2022automatic}
Farag, M.M., Fouad, M., Abdel-Hamid, A.T.: Automatic severity classification of diabetic retinopathy based on densenet and convolutional block attention module. IEEE Access  \textbf{10},  38299--38308 (2022)

\bibitem{guo2023visual}
Guo, M.H., Lu, C.Z., Liu, Z.N., Cheng, M.M., Hu, S.M.: Visual attention network. Computational Visual Media  \textbf{9}(4),  733--752 (2023)

\bibitem{howard2017mobilenets}
Howard, A.G., Zhu, M., Chen, B., Kalenichenko, D., Wang, W., Weyand, T., Andreetto, M., Adam, H.: Mobilenets: Efficient convolutional neural networks for mobile vision applications. arXiv preprint arXiv:1704.04861  (2017)

\bibitem{huang2023large}
Huang, T., Yin, L., Zhang, Z., Shen, L., Fang, M., Pechenizkiy, M., Wang, Z., Liu, S.: Are large kernels better teachers than transformers for convnets? In: International Conference on Machine Learning. pp. 14023--14038. PMLR (2023)

\bibitem{kim2023hrnet}
Kim, J.S., Park, S.W., Kim, J.Y., Park, J., Huh, J.H., Jung, S.H., Sim, C.B.: E-hrnet: Enhanced semantic segmentation using squeeze and excitation. Electronics  \textbf{12}(17), ~3619 (2023)

\bibitem{li2022dite}
Li, Q., Zhang, Z., Xiao, F., Zhang, F., Bhanu, B.: Dite-hrnet: Dynamic lightweight high-resolution network for human pose estimation. arXiv preprint arXiv:2204.10762  (2022)

\bibitem{li2020multi}
Li, X., Sun, S., Zhang, Z., Chen, Z.: Multi-scale grouped dense network for vvc intra coding. In: Proceedings of the IEEE/CVF Conference on Computer Vision and Pattern Recognition Workshops. pp. 158--159 (2020)

\bibitem{lin2014microsoft}
Lin, T.Y., Maire, M., Belongie, S., Hays, J., Perona, P., Ramanan, D., Doll{\'a}r, P., Zitnick, C.L.: Microsoft coco: Common objects in context. In: Computer Vision--ECCV 2014: 13th European Conference, Zurich, Switzerland, September 6-12, 2014, Proceedings, Part V 13. pp. 740--755. Springer (2014)

\bibitem{luo2022fastnet}
Luo, Y., Ou, Z., Wan, T., Guo, J.M.: Fastnet: Fast high-resolution network for human pose estimation. Image and Vision Computing  \textbf{119},  104390 (2022)

\bibitem{ma2018shufflenet}
Ma, N., Zhang, X., Zheng, H.T., Sun, J.: Shufflenet v2: Practical guidelines for efficient cnn architecture design. In: Proceedings of the European conference on computer vision (ECCV). pp. 116--131 (2018)

\bibitem{neff2021efficienthrnet}
Neff, C., Sheth, A., Furgurson, S., Middleton, J., Tabkhi, H.: Efficienthrnet: efficient and scalable high-resolution networks for real-time multi-person 2d human pose estimation. Journal of Real-Time Image Processing  \textbf{18}(4),  1037--1049 (2021)

\bibitem{ouyang2023efficient}
Ouyang, D., He, S., Zhang, G., Luo, M., Guo, H., Zhan, J., Huang, Z.: Efficient multi-scale attention module with cross-spatial learning. In: ICASSP 2023-2023 IEEE International Conference on Acoustics, Speech and Signal Processing (ICASSP). pp.~1--5. IEEE (2023)

\bibitem{rui2023edite}
Rui, L., Gao, Y., Ren, H.: Edite-hrnet: Enhanced dynamic lightweight high-resolution network for human pose estimation. IEEE Access  (2023)

\bibitem{sandler2018mobilenetv2}
Sandler, M., Howard, A., Zhu, M., Zhmoginov, A., Chen, L.C.: Mobilenetv2: Inverted residuals and linear bottlenecks. In: Proceedings of the IEEE conference on computer vision and pattern recognition. pp. 4510--4520 (2018)

\bibitem{sigal2021human}
Sigal, L.: Human pose estimation. In: Computer Vision: A Reference Guide, pp. 573--592. Springer (2021)

\bibitem{sun2019deep}
Sun, K., Xiao, B., Liu, D., Wang, J.: Deep high-resolution representation learning for human pose estimation. In: Proceedings of the IEEE/CVF conference on computer vision and pattern recognition. pp. 5693--5703 (2019)

\bibitem{tan2024object}
Tan, A., Guo, T., Zhao, Y., Wang, Y., Li, X.: Object detection based on polarization image fusion and grouped convolutional attention network. The Visual Computer  \textbf{40}(5),  3199--3215 (2024)

\bibitem{wang2021automated}
Wang, J., Qiao, X., Liu, C., Wang, X., Liu, Y., Yao, L., Zhang, H.: Automated ecg classification using a non-local convolutional block attention module. Computer Methods and Programs in Biomedicine  \textbf{203},  106006 (2021)

\bibitem{wang2020deep}
Wang, J., Sun, K., Cheng, T., Jiang, B., Deng, C., Zhao, Y., Liu, D., Mu, Y., Tan, M., Wang, X., et~al.: Deep high-resolution representation learning for visual recognition. IEEE transactions on pattern analysis and machine intelligence  \textbf{43}(10),  3349--3364 (2020)

\bibitem{wang2018non}
Wang, X., Girshick, R., Gupta, A., He, K.: Non-local neural networks. In: Proceedings of the IEEE conference on computer vision and pattern recognition. pp. 7794--7803 (2018)

\bibitem{wang2022lite}
Wang, Y., Li, M., Cai, H., Chen, W.M., Han, S.: Lite pose: Efficient architecture design for 2d human pose estimation. In: Proceedings of the IEEE/CVF Conference on Computer Vision and Pattern Recognition. pp. 13126--13136 (2022)

\bibitem{woo2018cbam}
Woo, S., Park, J., Lee, J.Y., Kweon, I.S.: Cbam: Convolutional block attention module. In: Proceedings of the European conference on computer vision (ECCV). pp. 3--19 (2018)

\bibitem{xiao2018simple}
Xiao, B., Wu, H., Wei, Y.: Simple baselines for human pose estimation and tracking. In: Proceedings of the European conference on computer vision (ECCV). pp. 466--481 (2018)

\bibitem{xu2022vitpose}
Xu, Y., Zhang, J., Zhang, Q., Tao, D.: Vitpose: Simple vision transformer baselines for human pose estimation. Advances in Neural Information Processing Systems  \textbf{35},  38571--38584 (2022)

\bibitem{yu2021lite}
Yu, C., Xiao, B., Gao, C., Yuan, L., Zhang, L., Sang, N., Wang, J.: Lite-hrnet: A lightweight high-resolution network. In: Proceedings of the IEEE/CVF conference on computer vision and pattern recognition. pp. 10440--10450 (2021)

\bibitem{yuan2021hrformer}
Yuan, Y., Fu, R., Huang, L., Lin, W., Zhang, C., Chen, X., Wang, J.: Hrformer: High-resolution transformer for dense prediction. arXiv preprint arXiv:2110.09408  (2021)

\bibitem{zhang2018shufflenet}
Zhang, X., Zhou, X., Lin, M., Sun, J.: Shufflenet: An extremely efficient convolutional neural network for mobile devices. In: Proceedings of the IEEE conference on computer vision and pattern recognition. pp. 6848--6856 (2018)

\bibitem{zhang2022efficient}
Zhang, X., Zeng, H., Guo, S., Zhang, L.: Efficient long-range attention network for image super-resolution. In: European conference on computer vision. pp. 649--667. Springer (2022)

\bibitem{zhang2022convolutional}
Zhang, Z., Wang, M.: Convolutional neural network with convolutional block attention module for finger vein recognition. arXiv preprint arXiv:2202.06673  (2022)

\bibitem{zheng2023deep}
Zheng, C., Wu, W., Chen, C., Yang, T., Zhu, S., Shen, J., Kehtarnavaz, N., Shah, M.: Deep learning-based human pose estimation: A survey. ACM Computing Surveys  \textbf{56}(1),  1--37 (2023)

\bibitem{zhou2022x}
Zhou, Y., Wang, X., Xu, X., Zhao, L., Song, J.: X-hrnet: Towards lightweight human pose estimation with spatially unidimensional self-attention. In: 2022 IEEE international conference on multimedia and expo (ICME). pp. 01--06. IEEE (2022)

\end{thebibliography}
\end{document}